\documentclass[10pt,twocolumn]{iopart}

\usepackage[pdftex]{graphicx}
\usepackage{iopams}
\usepackage{caption}
\usepackage{ulem}
\usepackage{xcolor}
\usepackage{dblfloatfix}
\expandafter\let\csname equation*\endcsname\relax

\expandafter\let\csname endequation*\endcsname\relax

\usepackage{amsmath}

\begin{document}

\title[Reservoir computing with large valid prediction time for the Lorenz system]{Reservoir computing with large valid prediction time for the Lorenz system}

\author{L A Hurley$^1$ and S E Shaheen$^{1,2,3}$}

\address{$^1$Department of Electrical, Computer and Energy Engineering, University of Colorado Boulder, Boulder, Colorado, 80309-0425 USA}
\address{$^2$Renewable and Sustainable Energy Institute, University of Colorado Boulder, Boulder, Colorado, 80309-0027, USA}
\address{$^3$Department of Physics, University of Colorado Boulder, Boulder, Colorado, 80309-0390, USA}
\ead{lauren.hurley@colorado.edu}
\vspace{10pt}
\begin{indented}
\item[]August 2025
\end{indented}



\begin{abstract}
We study the dependence of the Valid Prediction Time (VPT) of Reservoir Computers (RCs) on hyperparameters including the regularization coefficient, reservoir size, and spectral radius. Under carefully chosen conditions, the RC can achieve approximately 70\% of a benchmark performance, based on the output of a single prediction step used as initial conditions for the Lorenz equations. We report high VPT values ($>$30 Lyapunov times), as we are predicting a noiseless system where overfitting can be beneficial. While these conditions may not hold for noisy systems, they could still be useful for real-world applications with limited noise. Furthermore, utilizing knowledge of the Lyapunov exponent, we find that the VPT can be predicted by the error in the first few prediction steps, offering a computationally efficient evaluation method. We emphasize the importance of the numerical solver used to generate the Lorenz dataset and define a Valid Ground Truth Time (VGTT), during which the outputs of several common solvers agree. A VPT exceeding the VGTT is not meaningful, as a different solver could produce a different result. Lastly, we identify two spectral radius regimes that achieve large VPT: a small radius near zero, resulting in simple but stable operation, and a larger radius operating at the ``edge of chaos.''

\end{abstract}


%
\vspace{2pc}

\section{Introduction}

Reservoir Computing (RC) is a type of recurrent neural network consisting of fixed input layer weights, one hidden layer (the reservoir) with fixed weights, and an output layer whose weights are trained. One advantage of RC over other neural networks is that only the output layer weights are trained, resulting in fast and efficient training \cite{ESN, harnessing, modelf}. Although the output layer weights are the only parameter, there is a host of hyperparameters which must be carefully selected by the user. This includes the choice of activation function, input weight scaling, leakage rate, regularization parameter, reservoir size, sparsity, spectral radius, and weight distribution, among others \cite{global, proper}. The literature offers some guidance on choosing hyperparameters, such as the following publications: \cite{practical_guide, systematic_exploration, algorithms, analysis}. Still, there can be exceptions to these general guidelines depending on the task, and in those cases the choice of hyperparameters is often determined by trial and error. As this can be a time-consuming process, it may be neglected , and the operating conditions of the RC would not be fully optimized. 

Here, we investigate two hyperparameters in particular, the Ridge Regression coefficient $\lambda$ and the spectral radius, to optimize the Valid Prediction Time (VPT). A comprehensive search through radius and reservoir size at low $\lambda$ allows us to achieve higher VPTs than have been previously reported. We chose to investigate these two hyperparameters since they offer direct control over fundamental properties of the RC, such as dynamics, memory, and generalization \cite{practical_guide, systematic_exploration, noise}.

We find that the optimal $\lambda$ to optimize VPT is very small ($<10^{-20}$), which is nearly identical to a simple least squares regression. This trend agrees with existing literature, where it is generally considered that $\lambda$ is necessary for noisy systems but not for noiseless ones \cite{adaptive}. Moreover, preliminary work by dos Santos has noted that regularization may not be necessary depending on how the least squares problem is solved \cite{emergence}. They recommend that a good solver should not be sensitive to perturbations, have high numerical accuracy, be stable, and be robust to ill-conditioning. Yet, to our knowledge, no publications on RC predicting the Lorenz system have used this $\lambda$ dependence of VPT to their advantage, only reporting VPTs of around 15 \cite{systematic_exploration, variability}, while we report VPTs exceeding 30 Lyapunov times. 

Moreover, we find two regimes of spectral radius that lead to high VPT: one peak at small radii ($10^{-4}$) that does not depend on N, and another at higher radius which does depend on N, which we refer to as the edge of chaos (this may also be referred to as the edge of stability, see \cite{chaos}). The peak at the edge of chaos is slightly higher, but requires more precision in choice of N to attain, while the peak at low radius is an option that works for any N. Although, we note that the values slightly off-peak are still well above what has been reported in the literature. A spectral radius of zero means that the RC has no memory of past inputs and is an extreme learning machine \cite{extreme}. As the spectral radius is increased, the RC retains a memory of increasingly distant past inputs, but this comes at the cost of decreasing stability. Our findings agree with the literature, such as work by Jaurigue which found that that the highest VPT for a small reservoir (20 nodes) occurred with a radius close to zero \cite{benchmark}, and another by Platt et al. which found the best VPT for a large reservoir (2000 nodes) \cite{systematic_exploration}. Still, these are just individual cases, while the more thorough study of spectral radius and reservoir size presented here gives insight into previously hidden trends and regimes of performance.

Lastly, we emphasize the importance of careful choice of numerical solver methods. We find that the performance of the RC depends on the solver and tolerances used to generate the dataset. While there is some discussion of this in the literature (e.g. \cite{data_informed} notes that the sampling time of the solver used to generate the Lorenz system impacts the difficulty of the prediction task), we find that this tends to be overlooked, and we encourage the community to be specific in reporting the numerical methods used to generate chaotic time series datasets for RC and other neural network prediction tasks.

\section{Methods}\label{methods}

\subsection{Time Series Prediction}\label{background}

We simulate an echo state network (ESN) reservoir computer to predict a chaotic time series by training it on the history of that system's coordinates ${\bf z}$ sampled every $\tau$ time units during a training time $T$, $\{{\bf z}(n\tau)\}_{n=1}^{T/\tau}$. The system drives the internal states of the reservoir, ${\bf r}$, by using a fixed input matrix $W_{\text{in}}$ as
\begin{align}
{\bf r}((n+1)\tau) = f(A {\bf r}(n\tau) + W_{\text{in}} {\bf z}(n\tau)),\label{ridge}
\end{align}
where $A$ is the reservoir connectivity matrix with a spectral radius $\rho$ and $f(\cdot)$ is the activation function. The reservoir output is given by $\hat {\bf z}((n+1)\tau) = W_{\text{out}} {\bf r}((n+1)\tau)$, where the output matrix is chosen so that the output predicts as closely as possible the future state of the reservoir, i.e., $\textbf{z}((n+1)\tau)\approx \hat{\textbf{z}}((n+1)\tau)$. Here, we use a single least-squares regression that minimizes the cost function
\begin{align}
\sum_{n = 1}^{T/\tau}\| \hat {\bf z}(n\tau) - {\bf z}(n\tau)\|^2 + \lambda \text{Tr}(W_{\text{out}}  W_{\text{out}} ^T),
\end{align}

\noindent where $\lambda$ is a regularization term to prevent overfitting.

The Lorenz System, a system of three coupled differential equations, is a common benchmark for chaotic time series prediction and is given by \cite{Lorenz}

\begin{equation}\label{lorenz}
\begin{array}{l}
    \frac{dx}{dt} = \sigma(y-x),\\
    \frac{dy}{dt} = x(\rho -z) - y,\\
    \frac{dz}{dt} = xy - \beta z,
\end{array}
\end{equation}
with $\sigma = 10$, $\rho = 28$, and $\beta = 8/3$ (a.u.).

For prediction, the RC is trained on a numerical solution of Equation~(\ref{lorenz}), and then the output of the reservoir is used as the input for successive prediction steps. The RC prediction will align with the true system within a margin of error for a limited time. The time during which the prediction is accurate within a certain threshold can be quantified by metrics such as forecast horizon \cite{breaking_sym} or valid prediction time \cite{benchmark} and is further discussed in Sec.\ref{benchmark}.

\subsection{Simulations}\label{simulations}

As in our previous work, we use the Julia package ReservoirComputing.jl to simulate the reservoir computer \cite{tuning, julia}. Unless otherwise noted, the parameters used were as follows: the reservoir size (number of nodes) was $N = 300$. The reservoir matrix $A$ was constructed by choosing each entry to be $1$ with probability $6/N$ and $0$ otherwise, and then rescaling the matrix so that its spectral radius was $0.5$. The input matrix W$_\text{in}$ was chosen by assigning a random number uniformly distributed in $[-0.1,0.1]$ to each entry. The training method is a standard Ridge Regression with a regularization parameter $\lambda = 0$ [See Equation 2], the training time $T$ was 10000 and the prediction time was 5000. For the example shown in Figure 2, the activation function used was the {\it swish} function, $f(x) = x/(1+\exp(-\beta x))$ with $\beta = 1$.

\subsection{Metric: Valid Prediction Time}
To quantify the length of accurate prediction, we use the Valid Prediction Time (VPT) is defined as the time, in Lyapunov times, when the normalized squared difference between the prediction and the true data exceeds a threshold of 0.4 \cite{benchmark}. The normalized squared difference is defined as:

\begin{equation}
    E(t) = \frac{\sum_{i=1}^{D} (u_{\text{true},i}(t) - u_{\text{pred},i}(t))^2}{\text{Variance}(u_{\text{true}})}
\end{equation}

\noindent where $u_{\text{true},i}(t)$ is the true value of the $i$th dimension at time $t$, $u_{\text{pred},i}(t)$ is the predicted value of the $i$th dimension at time $t$, and $d$ is the number of dimensions. The denominator $\text{Variance}(u_{\text{true}})$ is given by:

\begin{equation}
    \text{Variance}(u_{\text{true}}) = \frac{1}{N} \sum_{t=1}^{N} \left( \sum_{i=1}^{D} \left( u_{\text{true},i}(t) - \frac{1}{N} \sum_{t=1}^{N} u_{\text{true},i}(t) \right)^2\right)
\end{equation}

\noindent where $N$ is the number of time steps. The VPT is determined by the first time $t$ where the normalized squared difference exceeds a threshold. Since the Lorenz attractor consists of two lobes, we note that the VPT threshold can vary without significant impacts on the result; that is, the VPT will occur when the RC ``chooses’’ the wrong side of attractor, and changing the threshold will hardly change the VPT.

\subsection{Benchmark}\label{benchmark}

As outlined in \cite{benchmark}, we define a ``best-case'' benchmark of RC performance. The RC is trained and used to predict one time step of the testing data. That prediction is then used as the initial conditions in the Lorenz system equations. This simulates the best possible performance of the RC with a given initial error. 

\section{Results and Discussion}

\subsection{Importance of Solver}
The solver used to generate the Lorenz data was ABM54 from the Julia package DifferentialEquations.jl, where the predictor is a fifth order Adams-Bashforth Explicit Method and a four-step Adams-Moulton is a corrector; the starting values are calculated with a fourth order Runge-Kutta \cite{ODE}. The solver used absolute and relative tolerances of $1\textsc{e}{-12}$ and a time step $\tau$ of $1\textsc{e}{-3}$. Under these conditions, the ABM54 solver agrees with a selection of other ODE solvers reasonably suited to the task (e.g. fourth order Runge-Kutta) up to around 30-40 Lyapunov times, which we term the Valid Ground Truth Time (VGTT) (see supplementary material Figures S1, S2, S3). Using smaller values for the absolute and relative tolerances and $\tau$ can increase the VGTT, but results in longer computation times. A VGTT of 40 Lyapunov times is shorter than a reasonable training dataset, which would range between 100-1000 Lyapunov times. Thus, when the RC is trained on Lorenz data generated by one solver, it might not necessarily have the same performance if its prediction were compared to the data generated by another solver. Moreover, the VGTT offers a computational upper-limit for the VPT: it is unreasonable to compare VPT-values higher than the VGTT because a different solver might result in a different solution.

\subsection{VPT Dependence on Regularization}

\begin{figure*}
  \begin{center}
  \includegraphics[width=0.8\linewidth]{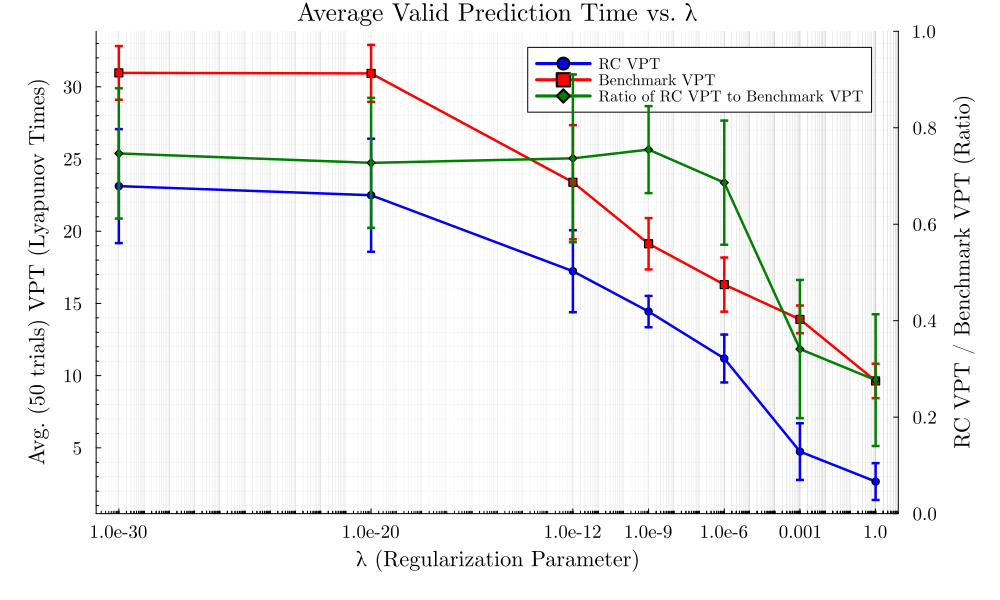}\\
  \caption{Valid Prediction Time, averaged over 50 trials, for the RC and the benchmark at different valus of $\lambda$, the ridge regression parameter. The benchmark performance depends on $\lambda$ and limits the possible VPT of the RC. The RC VPT is about 70\% of the benchmark for $0<\lambda<10^{-6}$ but then drops off. For comparison, the ratio of the RC VPT to benchmark VPT is plot on the right axis.}\label{fig1}
  \end{center}
\end{figure*}

Figure \ref{fig1} plots the dependence of the RC VPT, averaged over 50 trials, on $\lambda$, the Ridge Regression regularization parameter. Under the conditions outlined in Sec.~\ref{simulations}, the RC performs better at $\lambda$-values closer to zero. Previous literature has noted that non-zero $\lambda$ are necessary for training on noisy datasets \cite{RR}, but it is interesting to note that non-zero $\lambda$ are detrimental to VPT when training on a noise-free dataset. Moreover, we find that this trend is shared by the RC and the benchmark VPT, suggesting that the VPT is determined by the error of the RC in the first few prediction steps; this is further discussed in the following section. Lastly, we note that the experimental conditions used in this paper result in good performance, as they consistently result in a VPT which is about 70\% of the benchmark VPT for sufficiently small $\lambda$.

\subsection{VPT dependence on initial error}

\begin{figure*}
  \begin{center}
  \includegraphics[width=0.8\linewidth]{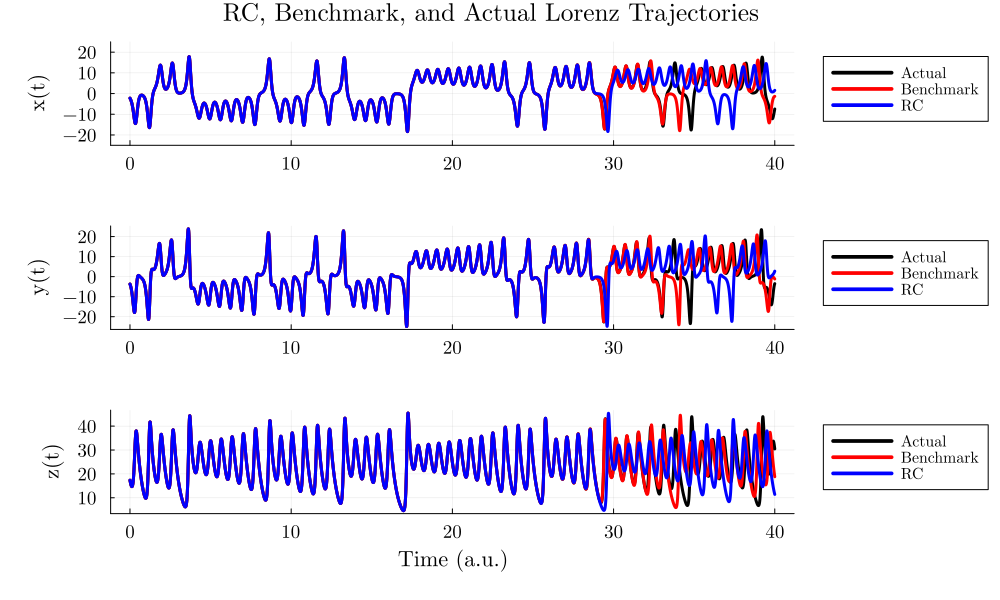}\\
  \caption{Trajectories for the Lorenz system, the benchmark, and the RC. (N.B that this was produced by a 400-node reservoir.) The horizontal axis is time in arbitrary units for easy comparison with Figure \ref{fig2b}. The benchmark accurately predicts the actual system slightly longer than the RC (about 5 Lyapunov times).}\label{fig2a}
  \end{center}
\end{figure*}

\begin{figure*}
  \begin{center}
  \includegraphics[width=0.8\linewidth]{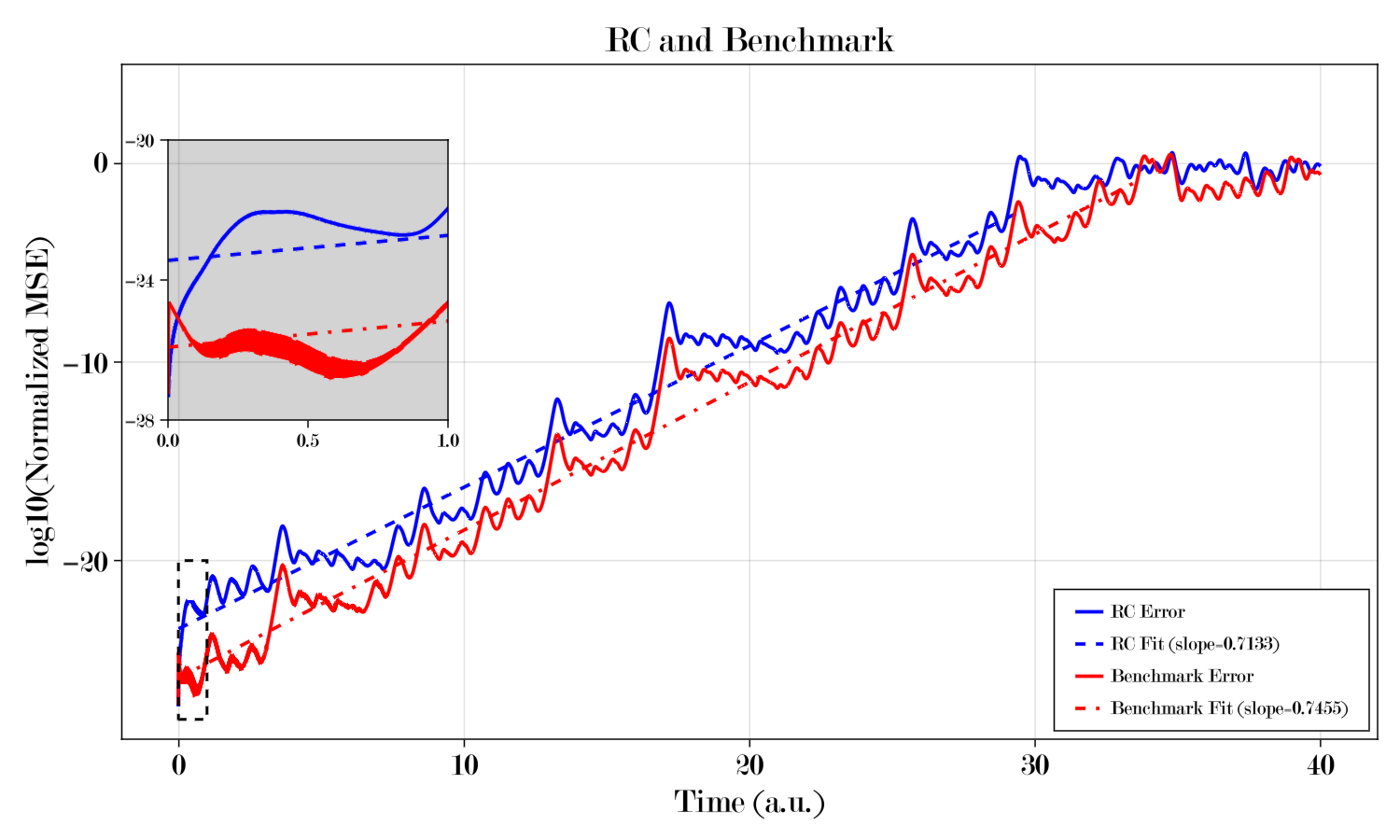}\\
  \caption{Normalized Mean Squared Error for the RC and the benchmark over time in arbitrary units. (N.B that this was produced by a 400-node reservoir.) The inset shows the evolution of error in the region outlined by the dashed rectangle on the main plot. Using time in arbitrary units allows for easy comparison with the slope of the fits with the Lyapunov exponent of the system. Here, the slopes are slightly less than the Lorenz system maximum Lyapunov exponent. Since the error evolves at nearly the same rate for both, an initial jump in error within the first few time steps is responsible for the reduced performance of the RC compared to the benchmark.}\label{fig2b}
  \end{center}
\end{figure*}

Figure \ref{fig2a} plots the evolution of the x, y, and z coordinates of the Lorenz system for (1) the results generated by the ODE solver, (2) the RC prediction, and (3) the benchmark prediction. Additionally, Figure \ref{fig2b} plots the evolution of the normalized mean squared error (MSE) on a logarithmic scale over time. Note that the horizontal axes of the plots in Figures \ref{fig2a}, \ref{fig2b} have arbitrary units, not Lyapunov time, so that the slope of the fit has meaning. Both the RC and the benchmark have nearly the same slope (0.7775 and 0.7580, respectively, with a percent difference of 2.54\%). Also note that this slope is close to but slightly less than the maximum Lyapunov exponent of the Lorenz system, 0.9056. Since the slopes of the RC and the benchmark are nearly identical, the VPT is therefore decided by the jump in error that occurs in the first few time steps. This can significantly reduce the amount of computation required to search for optimal hyperparameters, as it requires only the computation of the first few testing steps. For example, to assess different values of $\lambda$, one could choose the value with the smallest normalized MSE after a few testing steps instead of having to compute the output for all the testing steps and choosing the value with the longest VPT.

\subsection{VPT dependence on reservoir size and radius}

\begin{figure*}
  \begin{center}
  \includegraphics[width=0.8\linewidth]{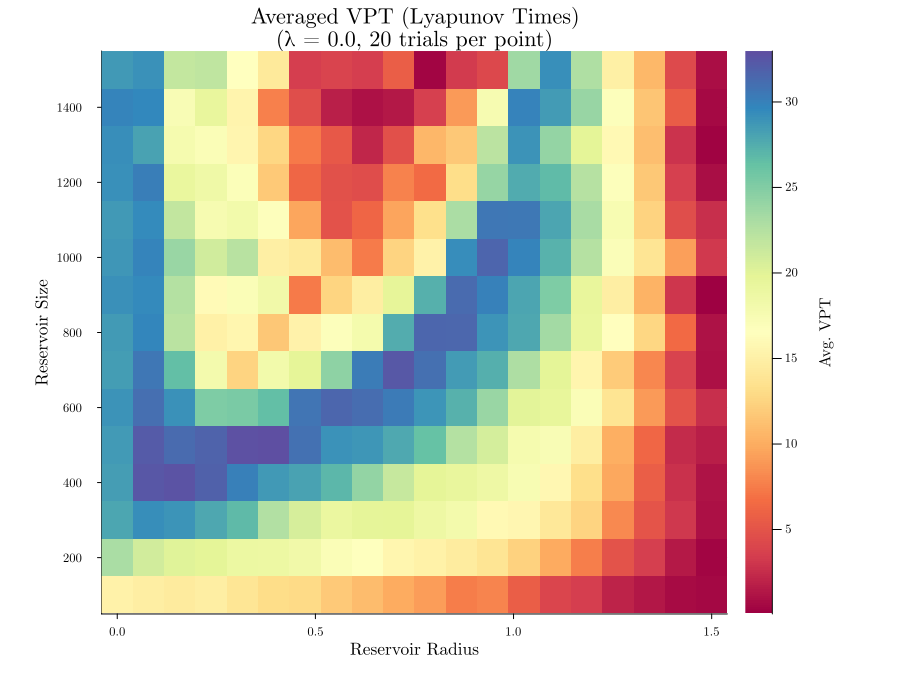}\\
  \caption{Valid Prediction Time averaged over 20 trials as a function of reservoir size (N) and radius. For a fixed N, there is more than one radius that results in optimal VPT. A small radius consistently results in high VPT regardless of N, but there is also a larger radius that will result in a high VPT and that depends on N.}\label{fig4}
  \end{center}
\end{figure*}

Previous work has investigated the relationship between spectral radius and performance for a fixed reservoir size \cite{benchmark, modelfree}. Here, we report a more detailed investigation of parameter space by plotting the average VPT as a function of both reservoir size and radius (Figure~\ref{fig4}). This reveals an interesting trend: there are two values of spectral radius that result in optimal VPT for a given N. A small radius (close to zero) performs consistently well across the N-values sampled. Even though a small radius means reduced memory and reservoir complexity, a sufficiently large N (here, N $> 300$) is able to compensate for that. Operating at low radius suggests a simpler operation mode that can still achieve good performance when the reservoir is large enough to offer sufficient degrees of freedom. We note that this agrees with the findings in \cite{benchmark}, which report an optimal VPT at small radius for N = 20.

Moreover, there is another peak VPT at a larger radius whose value depends on N. We consider this peak to occur at the ``edge of chaos,'' at the delicate balance between memory and stability. The fact that the optimal radius increases with increasing N suggests that larger N can compensate for a less stable system. 

\section{Conclusions}\label{conclusion}

In summary, we report an investigation of VPT dependence on select hyperparameters. We emphasize the importance of careful choice of numerical methods in generating data for RC training and testing. The time-step and tolerances of solvers must be sufficiently small for the output dataset to agree with the output of other solvers for a reasonable amount of time. If the reported VPT exceeds the time that the solvers agree (which we term the Valid Ground Truth Time), then the VPT is not strictly meaningful. We suggest that others working in this field be precise in reporting their numerical solver conditions.

Moreover, we find that to achieve high VPT, the optimal $\lambda$ is $\sim$ 0, to be expected of a noiseless model system in which strong overfitting is beneficial. To our knowledge, we report the largest VPT values ($>30$) to date for a RC predicting the Lorenz system. Furthermore, by exploring the evolution of MSE over time, we highlight that the VPT is determined by the error in the first few time steps. This can offer great reduction in the computational costs of exploring hyperparameter space since it only requires predicting the first few time steps. Lastly, we explore the dependence of VPT on reservoir size and radius. A small radius results in high VPT across reservoir size N, a stable regime where large enough N can compensate for lack of reservoir complexity. A larger radius whose value depends on N can also achieve high VPT, suggesting an ``edge of chaos'' regime where larger N can tolerate more chaotic reservoirs. Overall, this work reiterates the importance of deliberate hyperparameter selection for best performance and provides guidelines for achieving very large VPTs in the Lorenz system. We expect these conclusions to be qualitatively applicable to other noiseless, model dynamical systems; however, the details for other specific systems as well as the applicability to noisy, real-world time series are left to future work. 

\section*{Acknowledgments}

The authors would like to thank Prof. Juan G. Restrepo, Dept. of Applied Mathematics, for insightful discussions. We also thank the University of Colorado Boulder College of Engineering and Applied Science Interdisciplinary Research Theme (IRT) program and the Renewable and Sustainable Energy Institute (RASEI) for seed grant funds to support this work.

\section*{Data availability statement}
The data that support the findings of this study are available upon request from the authors.


\bibliographystyle{unsrt}

\end{document}


\title[Supplementary material: Reservoir computing with large valid prediction time for the Lorenz system]{Supplementary material: Reservoir computing with large valid prediction time for the Lorenz system}

\author{L A Hurley$^1$, and S E Shaheen$^{1,2,3}$}

\address{$^1$Department of Electrical, Computer and Energy Engineering, University of Colorado at Boulder}
\address{$^2$Renewable and Sustainable Energy Institute (RASEI)}
\address{$^3$Department of Physics, University of Colorado at Boulder}
\ead{lauren.hurley@colorado.edu}
\vspace{10pt}
\begin{indented}
\item[]August 2025
\end{indented}




%
%
%
%
%

\section{Simulation algorithm}
The algorithm used in the Echo State Network (ESN) simulation, with the Julia package ReservoirComputing.jl is as follows. For more information, please refer to the documentation for that package. For each trial,

\begin{enumerate}
    \item The Lorenz data is generated.
    \item The data is split into two sets, training and prediction. We used 5000 time units for the training and 1250 time units for prediction.
    \item The reservoir is generated according to the parameters specified in the main text.
    \item The output layer is trained using the training data and the specified training method as discussed in the main text.
    \item The Valid Prediction Time (VPT) is calculated using Equation 4 in the main text.
\end{enumerate}

At the end of all the trials, the VPT value is averaged.

\section{Valid Ground Truth Time (VGTT)}
\begin{figure*}
  \begin{center}
  \includegraphics[width=0.8\linewidth]{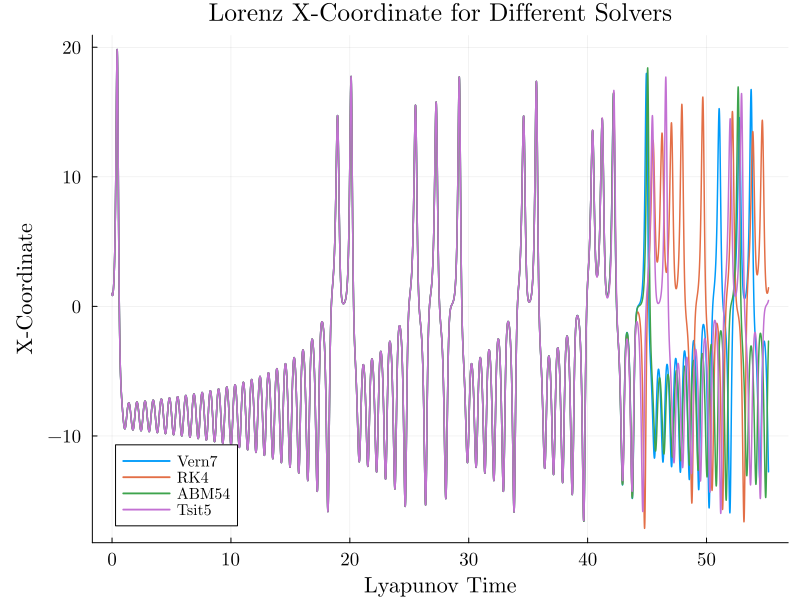}\\
  \caption{X-coordinate of the Lorenz system over time produced by a selection of solvers reasonably suited to the task. The solutions agree until about 43 Lyapunov times, which we term the VGTT.}\label{S1}
  \end{center}
\end{figure*}

\begin{figure*}
  \begin{center}
  \includegraphics[width=0.8\linewidth]{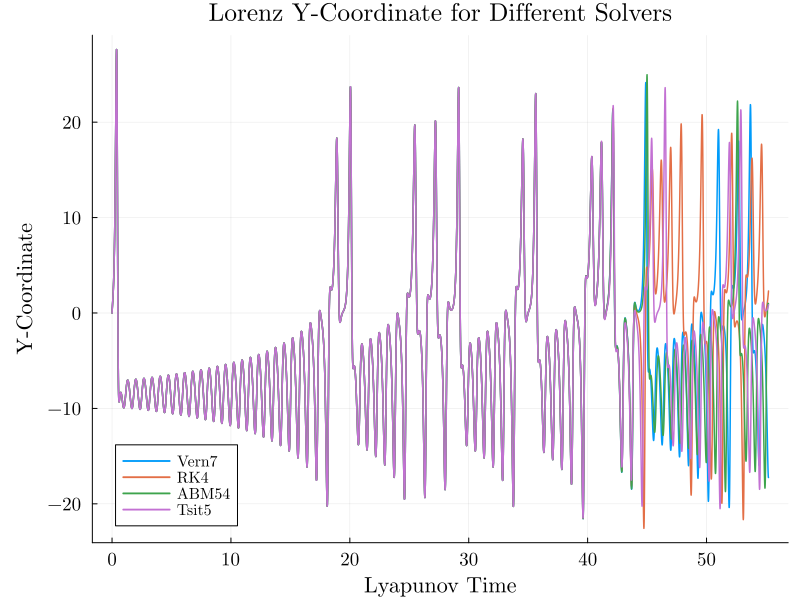}\\
  \caption{Y-coordinate of the Lorenz system over time produced by a selection of solvers reasonably suited to the task. The solutions agree until about 43 Lyapunov times, which we term the VGTT.}\label{S2}
  \end{center}
\end{figure*}

\begin{figure*}
  \begin{center}
  \includegraphics[width=0.8\linewidth]{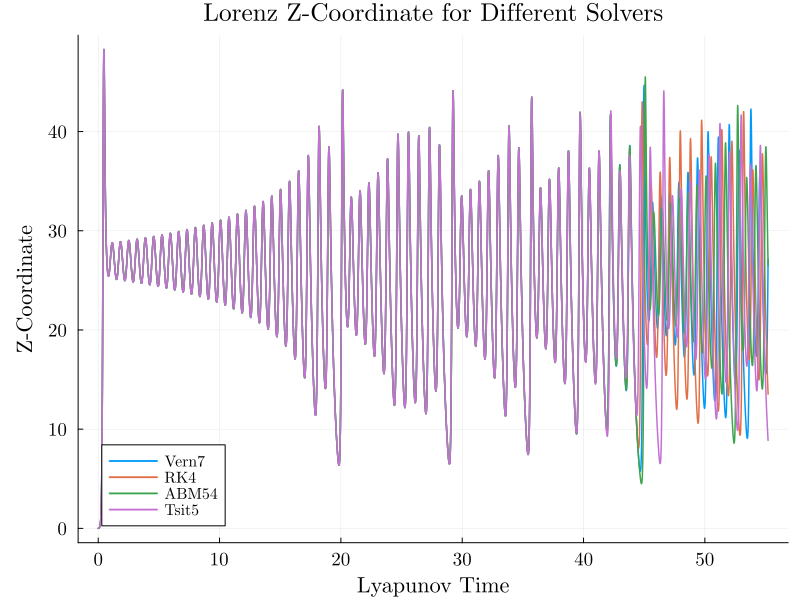}\\
  \caption{Z-coordinate of the Lorenz system over time produced by a selection of solvers reasonably suited to the task. The solutions agree until about 43 Lyapunov times, which we term the VGTT.}\label{S3}
  \end{center}
\end{figure*}

We define the Valid Ground Truth Time (VGTT) as the time during which the outputs from a selection of numerical solvers agree. This is illustrated in Figures \ref{S1}, \ref{S2}, \ref{S3}. To generate these graphs, the following conditions were used: a time step of $1e-5$, and absolute and relative tolerances of $1e-12$. The solvers used were from the Julia package DifferentialEquations.jl. Vern7 is Verner's “Most Efficient” 7/6 Runge-Kutta method. (lazy 7th order interpolant); RK4 is the canonical Runge-Kutta Order 4 method; ABM54 is s a fifth order Adams-Bashforth Explicit Method and a four-step Adams-Moulton is a corrector whose the starting values are calculated with a fourth order Runge-Kutta; and Tsit5 is Tsitouras 5/4 Runge-Kutta method (free 4th order interpolant). More information about these solvers can be found in the DifferentialEquations.jl documentation \cite{ODE}. 

\section{VPT Dependence on Initial Error}

\begin{figure*}
  \begin{center}
  \includegraphics[width=0.8\linewidth]{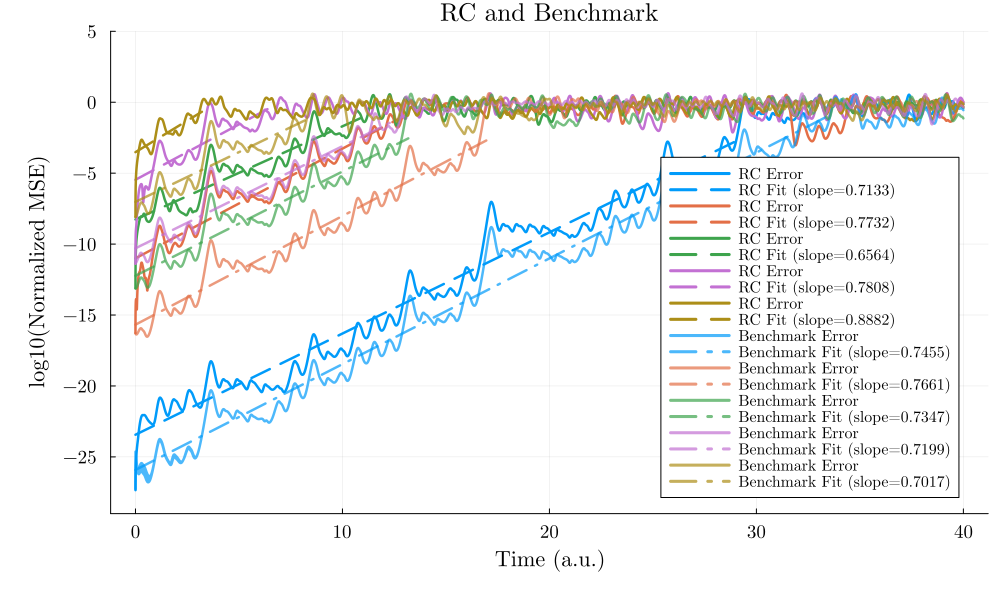}\\
  \caption{Normalized Mean Squared Error for the RC and the benchmark over time in arbitrary units, with different values of $\lambda$. The error in the first few time steps can be used to predict the VPT since the slope of the log of the NMSE is known.}\label{S4}
  \end{center}
\end{figure*}

Figure \ref{S4} is a version of Figure 3 in the main text but with multiple values of $\lambda$. This further illustrates how the error in the first few time steps plays a crucial role in determining the VPT.

\bibliographystyle{unsrt}